\documentclass[letterpaper, 10 pt, conference]{ieeeconf}  % Comment this line out if you need a4paper

\IEEEoverridecommandlockouts                              % This command is only needed if 
\usepackage{graphicx}
\usepackage{tikz}
\usepackage{comment}
\usepackage{amsmath,amssymb} % define this before the line numbering.
\usepackage{color}
\usepackage{cite}
\usepackage{pifont}
\usepackage{subcaption}
\usepackage{stix}
\usepackage{float}
\restylefloat{table}
\usepackage{placeins}

\title{\LARGE \bf Progressive Query Refinement Framework for Bird's-Eye-View Semantic Segmentation from Surrounding Images}

\author{Dooseop Choi$^{*1,2}$ and Jungyu Kang$^{1}$ and Taeghyun An$^{1}$ and Kyounghwan Ahn$^{1}$ and KyoungWook Min$^{1}$% <-this % stops a space
\thanks{*Corresponding author}% <-this % stops a space
\thanks{$^{1}$D. Choi, T. An, K. Ahn, and K. Min are with Superintelligence Creative Research Laboratory, ETRI, South Korea
        {\tt\small \{d1024.choi, tekkeni, mobileguru, kwmin92\}@etri.re.kr}}%
\thanks{$^{2}$D. Choi is Faculty of Artificial Intelligence, University of Science and Technology, South Korea}%
}

\begin{document}

\maketitle
\thispagestyle{empty}
\pagestyle{empty}

%%%%%%%%%%%%%%%%%%%%%%%%%%%%%%%%%%%%%%%%%%%%%%%%%%%%%%%%%%%%%%%%%%%%%%%%%%%%%%%%
\begin{abstract}

Expressing images with Multi-Resolution (MR) features has been widely adopted in many computer vision tasks. In this paper, we introduce the MR concept into Bird's-Eye-View (BEV) semantic segmentation for autonomous driving. This introduction enhances our model's ability to capture both global and local characteristics of driving scenes through our proposed residual learning. Specifically, given a set of MR BEV query maps, the lowest resolution query map is initially updated using a View Transformation (VT) encoder. This updated query map is then upscaled and merged with a higher resolution query map to undergo further updates in a subsequent VT encoder. This process is repeated until the resolution of the updated query map reaches the target. Finally, the lowest resolution map is added to the target resolution to generate the final query map. During training, we enforce both the lowest and final query maps to align with the ground-truth BEV semantic map to help our model effectively capture the global and local characteristics. We also propose a visual feature interaction network that promotes interactions between features across images and across feature levels, thus highly contributing to the performance improvement. We evaluate our model on a large-scale real-world dataset. The experimental results show that our model outperforms the SOTA models in terms of IoU metric. Codes are available at https://github.com/d1024choi/ProgressiveQueryRefineNet

\end{abstract}

%%%%%%%%%%%%%%%%%%%%%%%%%%%%%%%%%%%%%%%%%%%%%%%%%%%%%%%%%%%%%%%%%%%%%%%%%%%%%%%%
\section{INTRODUCTION}

Perceiving driving environments from surrounding camera images has recently gained significant attention in autonomous driving. This is because camera sensors are not only typically more cost-effective than other sensors, such as LIDAR and RADAR, but also provide rich semantic information that other sensors cannot capture. Detecting 3D objects from the images is one of the actively researched tasks. Another promising task that leverages the abundant information in the images is predicting BEV semantic maps of driving scenes with respect to autonomous vehicles. Since autonomous driving is inherently a geometric problem, where the goal is to navigate a vehicle safely and correctly through 3D space \cite{Hu_iccv21, Choi}, BEV maps can be directly deployed for the subsequent tasks such as motion planning and control. 

% (removed) These maps can represent not only surrounding objects (e.g., vehicles, pedestrians) but also the geometric layouts of driving environments, including drivable space and lane lines, from an orthographic BEV perspective.

Presenting input images at multiple resolutions has long been a common practice in various computer vision tasks. This allows NN models to capture both local and global characteristics of the images effectively. One pioneering work is the Feature Pyramid Network (FPN) \cite{lin_cvpr17}, which is proposed for the 2D object detection task, and the majority of the models dedicated to the VT leverage the FPN to better capture objects of different sizes in driving images. BEV semantic maps of driving scenes, which typically are created by rendering HD map components and detected 3D objects on a 2D canvas, also possess their own set of global and local characteristics. For instance, 3D objects are depicted as rectangles with varying sizes on the map, whereas lane lines are represented as lines of different lengths. These characteristics can be effectively conveyed when the maps are represented at multiple resolutions.

% (removed) One pioneering work is the Feature Pyramid Network (FPN) \cite{lin_cvpr17} proposed for 2D object detection, in which MR image features are scaled and merged to better capture both large and small objects in images.

In this paper, we propose a VT model that leverages MR BEV query maps to better capture the global and local characteristics of driving scenes. In particular, given randomly generated MR BEV query maps, the lowest resolution map is initially updated by a VT encoder. The updated map is then upscaled and merged with a higher resolution one to be fed into a subsequent VT encoder for further updates. This process repeats until the updated query map reaches the target resolution. The final query map is then obtained by adding the lowest resolution to the target resolution. We supervise both the lowest and final maps during training. As a consequence, the VT encoder dedicated to the lowest resolution can learn to capture the global characteristics while the other VT encoders dedicated to higher resolution ones contribute to capturing missing details. We also propose a visual feature interaction network to promote the interaction between features across images and feature levels, which in turn contributes to the performance improvement.

\section{Related Works}
$\textbf{View Transformation:}$ Transforming features in Camera View (CV) to BEV is considered an ill-posed problem, primarily because camera depth information is generally assumed to be unknown. To address this issue, some early works proposed training NNs to directly transform image features to features in BEV space without considering the camera pose and depth \cite{Roddick_cvpr20, Pan_ral20}. LSS \cite{Philion_eccv20} is the first to propose predicting the depth information for the forward projection, where image features are projected into BEV space according to the predicted depth and camera intrinsic/extrinsic parameters. Based upon the forward projection paradigm, subsequent works proposed methods that better capture the BEV representation for tasks such as BEV segmentation \cite{Hu_iccv21, Hu_eccv22, Liu_icra23, Zhu_iccv23}, motion prediction and planning \cite{Hu_iccv21, Hu_eccv22, Akan_eccv22}, and 3D object detection \cite{Li_iccv23, Liu_icra23}. 

\begin{figure}[t]
\centering
\includegraphics[height=4.0cm]{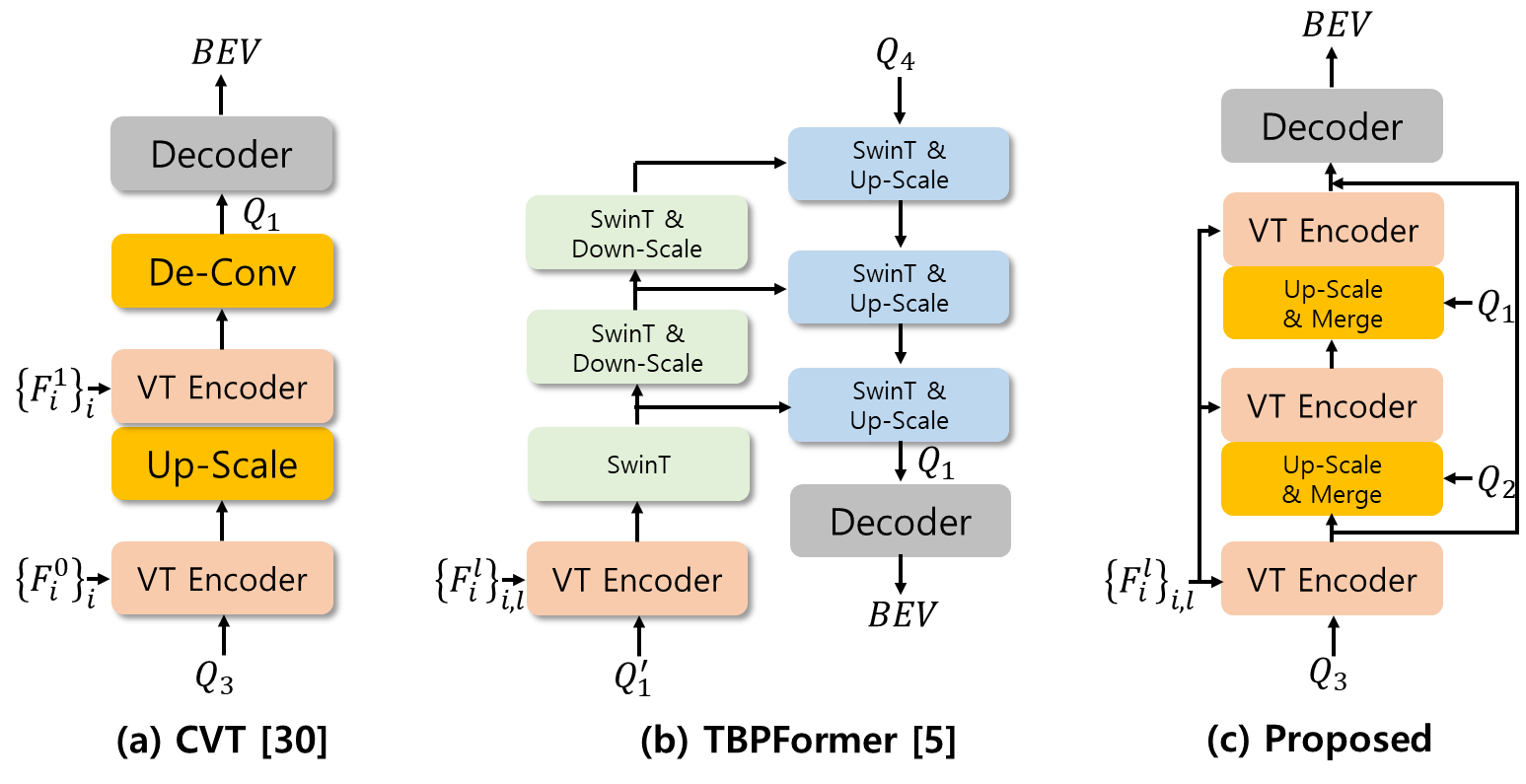}
\caption{Different progressive query refinement methods. (c) briefly depicts the proposed architecture.}
\label{fig1}
\end{figure}

The backward projection paradigm projects 3D reference points in BEV space to image plane to sample image features and back-project them to the BEV grid. Consequently, it avoids the additional operations for the depth prediction and mitigates the sparse mapping problem in the forward projection. OFT \cite{Roddick_bmvc19} is the first to propose using the backward projection for the VT. Building upon this, BEVFormer \cite{Li_eccv22} further suggested leveraging the powerful capability of the deformable attention \cite{Zhu_iclr21} for learning the VT. In accordance with the backward projection paradigm, subsequent works proposing improved performance have been introduced for various tasks such as BEV segmentation \cite{Fang_cvpr23}, 3D segmentation \cite{Huang_cvpr23}, and 3D object detection \cite{Yang_cvpr23, Wang_cvpr23}. In this paper, we deploy the VT encoder proposed in BEVFormer \cite{Li_eccv22} for the BEV query map update. However, as illustrated in Fig. \ref{fig1}, any query-based VT encoder can be easily deployed within our framework.

% (removed) The explicit transformation paradigms mentioned above can be inflexible and create a rigid bottleneck during training and inference \cite{Zhou_cvpr22}. To address this issue, some authors proposed methods that allow NNs to learn the correspondence between image and BEV features directly through cross-attention operations \cite{Vaswani_nips17}. They also suggested encoding 3D position information \cite{Liu_eccv22, Liu_iccv23, Xiong_cvpr23} or the intrinsic/extrinsic parameters \cite{Zhou_cvpr22, Pan_cvpr23} into the image features and BEV queries, enabling NNs to learn the correspondence more accurately.

% Fig 1------------------------------------
% Fig 1------------------------------------
$\textbf{Multi-Resolution BEV Query Representation:}$ Representing the BEV space with MR query maps has been adopted in preceding works \cite{Zhou_cvpr22, Pan_cvpr23, Liu_eccv22, Li_iccv23}, primarily for the reduction of computational complexity. Specifically, as depicted in Figure \ref{fig1}a, an LR BEV query map is updated using a VT encoder and then upscaled for further updates in a subsequent VT module. This process is repeated until the resolution reaches an affordable level. The BEV feature map of the target resolution is then finally obtained via an up-sampling network. TBPFormer \cite{Fang_cvpr23} proposed first updating the target resolution query map via BEVFormer \cite{Li_eccv22} and then producing MR query maps from the target resolution map through SwinT \cite{Liu_iccv21} and downsampling operations. The generated MR query maps are then utilized to progressively update and upsample another LR query map up to the final resolution as depicted in Figure \ref{fig1}b. 

In this paper, we begin with a set of randomly generated MR query maps. These are progressively updated and merged starting from the lowest resolution to the target resolution as depicted in Figure \ref{fig1}c. The final query map is then obtained by adding the lowest resolution to the target resolution. While the existing works supervise only the target resolution during training, we propose supervising both the lowest and final maps. Supervising the lowest resolution map during the training forces the VT encoders dedicated to the higher resolution maps to learn missing details in the lowest map, which can be regarded as the residual learning \cite{He_cvpr16}.

$\textbf{Visual Feature Interaction}:$ Objects and backgrounds in an image tend to have specific relationships within a context. For example, pedestrians tend to be on sidewalks, while vehicles are on roads. In order to capture these relationships and leverage them for the tasks at hand, many works have proposed promoting interactions between features in an image through attention mechanism \cite{Dosovitskiy_iclr20, Carion_eccv20, Liu_iccv21, Zhu_iclr21, Xia_cvpr22}. For the VT task, however, less attention has been devoted to promoting the interactions primarily because BEV queries can easily attend to relevant regions across different images through the attention, and interactions arise implicitly during the attention process. Peng et al. \cite{Peng_wacv23} proposed promoting interactions between features in an image at different levels through deformable attention \cite{Zhu_iclr21}. Pan et al. \cite{Pan_cvpr23} attempted to promote the interactions implicitly through their proposed bi-direction cross-attention in which image features are updated from the refined BEV query maps. In this paper, we propose a visual feature interaction network designed to promote interactions between features not only across feature levels but also across images.

\section{Proposed Approach}
\subsection{Problem Formulation}
Suppose that an autonomous vehicle (AV) is equipped with $N_{c}$ surrounding cameras. The $i$-th camera image acquired at time $t$ is denoted as $I_{i,t} \in \mathbb{R}^{H_{I} \times W_{I} \times 3}$, where $H_{I}$ and $W_{I}$ denote the height and width of the image, respectively. Our target is to obtain a BEV feature map $\mathbf{B}_{t}\in \mathbb{R}^{X \times Y \times C}$, which is utilized as input to subsequent NNs for the semantic BEV map prediction, from the images $\{ I_{i,t} \}_{i=1}^{N_{c}}$. Here, $X$ and $Y$ respectively denote the length and width of the BEV grid space spanned by the vehicle pose at $t$, and $C$ denotes the BEV feature dimension. In the rest of this paper, we omit $t$ for the sake of simplicity. Our proposed VT module takes a set of MR BEV query maps $\{ \mathbf{Q}_{s} \}_{s=1}^{N_{s}}$ and MR image feature maps $\{ \mathbf{F}_{i}^{l} \}_{i=1,l=1}^{N_{c}, N_{l}}$ as input and produces $\mathbf{B}$ by updating and merging the query maps progressively. Here, $N_{s}$ is the number of the MR query maps and $\mathbf{Q}_{s} \in \mathbb{R}^{\frac{X}{2^{s-1}} \times \frac{Y}{2^{s-1}} \times C}$. The image feature map $\mathbf{F}_{i}^{l}$ is the $l$-th layer output of an image backbone (e.g., ResNet \cite{He_cvpr16}) when $I_{i}$ is used as input.

% Fig 2------------------------------------
\begin{figure*}[t]
\centering
\includegraphics[height=5.0cm]{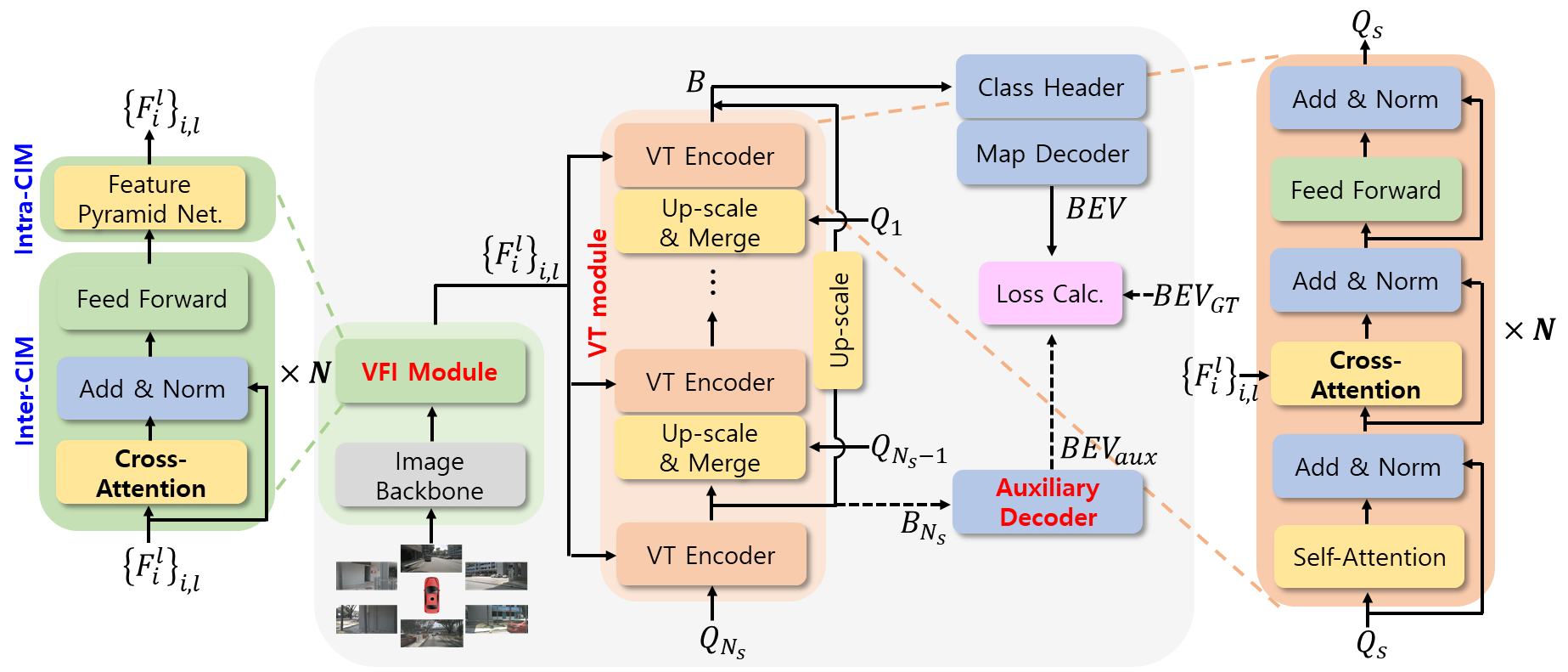}
\caption{The overall architecture of the proposed segmentation network}
\label{fig2}
\end{figure*}
% Fig 2------------------------------------

\subsection{Network Architecture}
Figure \ref{fig2} briefly illustrates the overall architecture of our proposed BEV segmentation network. First, MR feature maps $\{ \mathbf{F}_{i}^{l} \}$ are extracted from $\{ I_{i} \}$ using a backbone network. The proposed visual feature interaction (VFI) module is then applied to the MR feature maps to promote interactions between the features both across levels and across images. Next, the proposed VT module, which consists of VT encoders and upsamplers, updates and merges $\{ \mathbf{Q}_{s} \}$ progressively to produce $\mathbf{B} = \mathbf{Q}_{1} + \mathbb{UpScale}(\mathbf{Q}_{N_{s}}) \in \mathbb{R}^{X \times Y \times C}$. Finally, a class header network, which is dedicated to a specific object or road element class, is applied to $\mathbf{B}$ and the final BEV map $\mathbf{BEV} \in \mathbb{R}^{X \times Y}$ is predicted using a subsequent map decoder network. As we mentioned, supervising both $\mathbf{B}$ and $\mathbf{Q}_{N_{s}}$ during training can be regarded as residual learning \cite{He_cvpr16}, resulting in a significant improvement in segmentation performance.

% Fig 3------------------------------------
\begin{figure}[t]
\centering
\includegraphics[height=3.0cm]{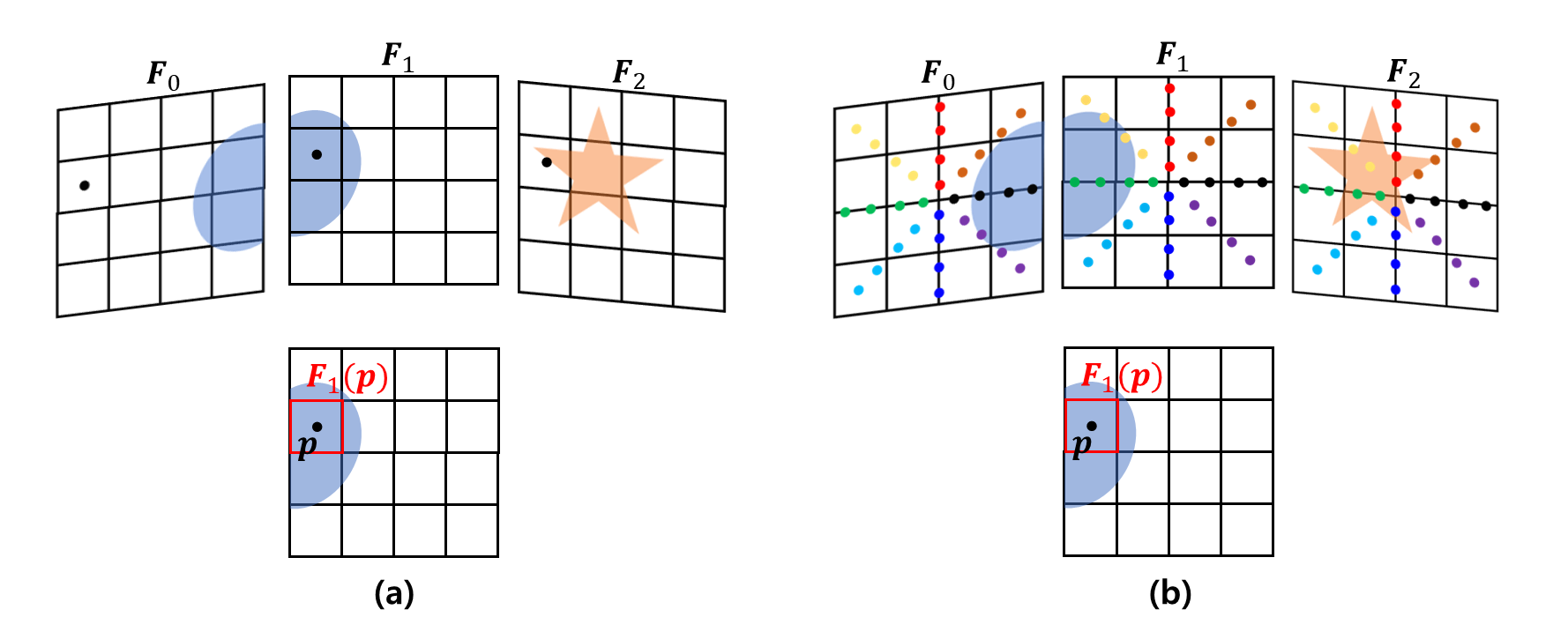}
\caption{Examples of (a) the conventional and (b) the proposed 2D reference points patterns. }
\label{fig3}
\end{figure}
% Fig 3------------------------------------

$\textbf{Visual feature interaction (VFI) module}$ is devised to promote interactions between the MR image features. It comprises two sub-modules: 1) Intra-camera interaction module (Intra-CIM), 2) Inter-camera interaction module (Inter-CIM). Intra-CIM is introduced to promote interactions between features located at the same pixel position across different feature levels and is implemented by the FPN \cite{lin_cvpr17}. Inter-CIM is introduced to promote interactions between features across different camera images. Given a set of image features $\textit{at the same feature level}$, $\{ \mathbf{F}_{i}\}$, the interactions are promoted through our proposed multi-head deformable attention as follows:
\begin{equation}
f_{DA}(\mathbf{F}_{i}(\mathbf{p}), \{ \mathbf{F}_{j}\}_{j=1}^{N_{c}}) =\sum_{h \in \mathcal{H}} \sum_{\mathbf{p} \in \mathcal{P}_{h}} \sum_{i=1}^{N_{c}} \alpha_{\mathbf{p}, i}^{h} \mathbf{F}_{i}(\mathbf{p} + \Delta \mathbf{p}_{i}^{h}),
\label{eqn1}
\end{equation}
where $\mathcal{H}$ is a set of head indices, $\mathcal{P}_{h}$ is a set of 2D reference points for a head of index $h$, and $\alpha_{\mathbf{p}, i}^{h}$ denotes an attention weight and satisfies $\sum_{\mathbf{p}, i} \alpha_{\mathbf{p}, i}^{h} = 1$. Our deformable attention distinguishes itself from the original \cite{Zhu_iclr21} through the design of the reference points pattern and the calculation of offsets $\{\Delta \mathbf{p}_{i}^{h} \}$ and weights $\{ \alpha_{\mathbf{p}, i}^{h} \}$. Let us elaborate on our reference points pattern using Fig. \ref{fig3}b. For the deformable attention operation, we assign each head four reference points, covering specific areas in an image. We illustrate in Fig. \ref{fig3}b the reference points for each head with different colors. As a consequence, with the eight heads having distinct reference points, which cover different image areas, our model can easily locate the image regions to be attended to. On the other hand, in the conventional design, the reference points for each head are the same as the pixel position of the query feature, $\mathbf{p}$, as seen in Fig. \ref{fig3}a. 

% ------------------------------------------------------------ Table 1
\begin{table*}[t]
\begin{center}
\caption{Experiments on nuScenes dataset. The values in this table are from the corresponding papers. * denotes that values are estimated from the respective paper assuming no data augmentation during training. $\textit{s}$ stands for $\textit{static}$, indicating that only current-time image data is used for prediction. We exclude the result of ST-P3 \cite{Hu_eccv22} for \textit{Lane Lines} since it didn't follow the common practice for the ground-truth BEV semantic map generation.}\label{table1}
\begin{tabular}{|c|c|c|c|c|c|}
\hline
Model & Input Size & Vehicle & Pedestrian & Drivable Area & Lane Lines \\
\hline
CVT \cite{Zhou_cvpr22} & 224 $\times$ 480 & 36.0 & - & 74.3 & - \\
LSS \cite{Philion_eccv20} & 224 $\times$ 480 & 32.1 & 15.0 & 72.9 & 20.0 \\
FIERY-\textit{s} \cite{Hu_iccv21} & 224 $\times$ 480 & 35.8 & - & - & - \\
ST-P3 \cite{Hu_eccv22} & 224 $\times$ 480 & 40.1 & 14.5 & 75.9 & - \\
TBPFormer-\textit{s}* \cite{Fang_cvpr23} & 224 $\times$ 480 & 43.6 & $\mathbf{16.1}$ & - & - \\
BAEFormer \cite{Fang_cvpr23} & 224 $\times$ 480 & 38.9 & - & 76.0 & - \\
BEVFormer-\textit{s} \cite{Li_eccv22} & 900 $\times$ 1600 & 43.2 & - & 80.7 & 21.3 \\
\hline
Ours & 224 $\times$ 480 & $\mathbf{43.7}$ & 15.7 & $\mathbf{81.1}$ & $\mathbf{25.6}$ \\
\hline
\end{tabular}
\end{center}
\end{table*}
% ------------------------------------------------------------

In the conventional deformable attention \cite{Zhu_iclr21}, the offsets and weights are predicted directly from $\mathbf{q}_{\textbf{emb}}=\mathbf{F}_{i}(\mathbf{p})+\mathbf{P}(\mathbf{p})$ through MLPs, where $\mathbf{P}(\mathbf{p})$ denotes a sinusoidal positional embedding for the pixel position $\mathbf{p}$. We, in this paper, predict the offsets and weights as proposed in \cite{Zhu_iclr21} with the following two modifications: 1) We introduce trainable embedding vectors 
$\{ \mathbf{c}_{i} \}_{i=1}^{N_{c}}$ to encourage the VFI module to distinguish image features across different images and to enhance the prediction of the offsets and weights. As a result, the offsets and weights are predicted from $\mathbf{q}_{\textbf{emb}}=\mathbf{F}_{i}(\mathbf{p})+\mathbf{P}(\mathbf{p})+\mathbf{c}_{i}$ through MLPs. 2) We limit the maximum allowable magnitude of the offsets to ensure that a point $\mathbf{p} + \Delta \mathbf{p}_{i}^{h}$ can cover a specific area of an image as follows:
\begin{equation}
\Delta \mathbf{p}_{i}^{h} = \delta \cdot \text{Tanh} ( \text{MLP}(\mathbf{q}_{\textbf{emb}})),
\label{eqn2}
\end{equation}
where $\text{Tanh}()$ denotes the hyperbolic tangent function. $\delta$ is a positive constant and is set to 0.25 in this paper.

\textbf{View Transformation (VT) Encoder} repeatedly updates $\mathbf{Q}_{s}$ using $\{ \mathbf{F}_{i}^{l} \}$. We deploy the Transformer encoder proposed in \cite{Li_eccv22}, which benefits from the deformable attention \cite{Zhu_iclr21} in terms of complexity and training speed. Specifically, let $\mathbf{V}$ be the projection of $\mathbf{Q}_{s}$ through a MLP. Then, the self-attention module in the VT encoder updates $\mathbf{Q}_{s}$ via the following deformable attention:
\begin{equation}
f_{DA}(\mathbf{Q}_{s}(\mathbf{p}), \mathbf{V}) =\sum_{h \in \mathcal{H}} \sum_{z=1}^{Z} \alpha_{z}^{h} \mathbf{V}(\mathbf{p} + \Delta \mathbf{p}_{z}^{h}),
\label{eqn3}
\end{equation}
where $\mathbf{Q}_{s}(\mathbf{p})$ denotes the element of $\mathbf{Q}_{s}$ at the spatial location $\mathbf{p}$ in the BEV grid space. $Z$ is the number of the value sampling operation. By adding the result of (3) to $\mathbf{Q}_{s}(\mathbf{p})$ and applying a feed forward network (FFN) to the addition, the interaction between queries in $\mathbf{Q}_{s}$ is promoted.

% where $\mathbf{Q}_{s}(\mathbf{p})$ denotes the element of $\mathbf{Q}_{s}$ at the spatial location $\mathbf{p} = [x, y]^{T}$, $0 \leq (x, y) \leq (X, Y)$ in the BEV grid space.

% (removed) The attention weight $\alpha_{z}^{h}$ and offset $\Delta \mathbf{p}_{z}^{h}$ are predicted directly from $\mathbf{Q}_{s}(\mathbf{p})$ through MLPs and $\sum_{z=1}^{Z} \alpha_{z}^{h} = 1$ holds.

The cross-attention module in the VT encoder updates $\mathbf{Q}_{s}$ using the visual features $\{ \mathbf{F}_{i}^{l} \}$ through the deformable attention defined as follows:
\begin{multline}
f_{DA}^{h}(\mathbf{Q}_{s}(\mathbf{p}), \{ \mathbf{F}_{i}^{l} \}, \{\mathcal{T}_{i}\}) \\ 
= \frac{1}{|\mathcal{V}_{hit}|} \sum_{i \in \mathcal{V}_{hit}} \sum_{z=1}^{Z} \sum_{l=1}^{N_{l}}
\alpha_{z,l}^{i} \mathbf{V}_{i}^{l}(\mathcal{T}_{i}(\mathbf{p}, i, z) + \Delta \mathbf{p}_{z,l}^{i}),
\label{eqn4}
\end{multline}
where $f_{DA}^{h}$ denotes the deformable attention dedicated to the $h$-th header and $\mathbf{V}_{i}^{l}$ is the projection of $\mathbf{F}_{i}^{l}$ through a MLP. $\mathcal{T}_{i}$ is the transformation function that projects a 3D point in the BEV space into a 2D point in the $i$-th image space. Consequently, $\mathcal{T}_{i}(\mathbf{p},i,z)$ is a 2D point in the $i$-th image space projected from the $z$-th 3D reference point defined on the position $\mathbf{p}$ in the BEV space. $\mathcal{V}_{hit}$ is the set of image indices where at least one of the 3D reference points is projected to. Finally, we note that the attention weight $\alpha$ and offset $\Delta \mathbf{p}$ are predicted directly from $\mathbf{Q}_{s}(\mathbf{p})$ through MLPs after a positional sinusoidal embedding is added to $\mathbf{Q}_{s}(\mathbf{p})$.

\textbf{Query Map Merge Module} fuses a previously updated LR query map $\mathbf{Q}_{s-1}$ with a higher resolution query map $\mathbf{Q}_{s}$ for a subsequent query update. We explored diverse options for the fusion and empirically found that a simple addition operation yields the best performance. Specifically, $\mathbf{Q}_{s-1}$ first undergoes bilinear interpolation to match the size of $\mathbf{Q}_{s}$ and then is added to $\mathbf{Q}_{s}$. Finally, $\mathbf{Q}_{s}$ goes through a subsequent VT encoder.

\textbf{Auxiliary Task} is introduced to encourage our model to capture the global characteristics of driving scene more effectively. Our auxiliary decoder depicted in Figure \ref{fig2} progressively increases the resolution of the lowest resolution query map up to the target resolution to obtain the final output $\mathbf{BEV}_{aux}\in \mathbb{R}^{X \times Y}$. It comprises a series of NN modules, each of which consists of Up-scaler, Conv, BN \cite{Ioffe_icml15}, and Relu. During the training, $\mathbf{BEV}_{aux}$ is forced to match the ground-truth BEV semantic map, $\mathbf{BEV}_{GT}$. As shown in the section IV, our model achieves the best performance when only the lowest resolution query map undergoes the auxiliary task.

% Here, $N_{dc}$ denotes the number of different classes to be considered for the segmentation.

\subsection{Losses}
To train our model, we minimize the final loss $\mathcal{L}$ defined as follows:
\begin{equation}
\mathcal{L} = \sum_{c \in \mathcal{C}} \lambda_{c} (\mathcal{L}_{main}^{c} + \alpha \mathcal{L}_{aux}^{c}),
\label{eqn5}
\end{equation}
where $\mathcal{L}_{main}^{c}$ and $\mathcal{L}_{aux}^{c}$ denote the focal losses \cite{Lin_iccv17} for a specific class $c$ calculated from $\mathbf{BEV}$, $\mathbf{BEV}_{aux}$, and $\mathbf{BEV}_{GT}$. $\alpha$ and $\lambda_{c}$ are pre-defined constants. $\alpha$ balances the contributions of $\mathcal{L}_{main}$ and $\mathcal{L}_{aux}$ to the final loss. $\lambda_{c}$ balances the contributions of each class to the final loss.

% Fig 4------------------------------------
\begin{figure*}[t]
\centering
\includegraphics[height=6.0cm]{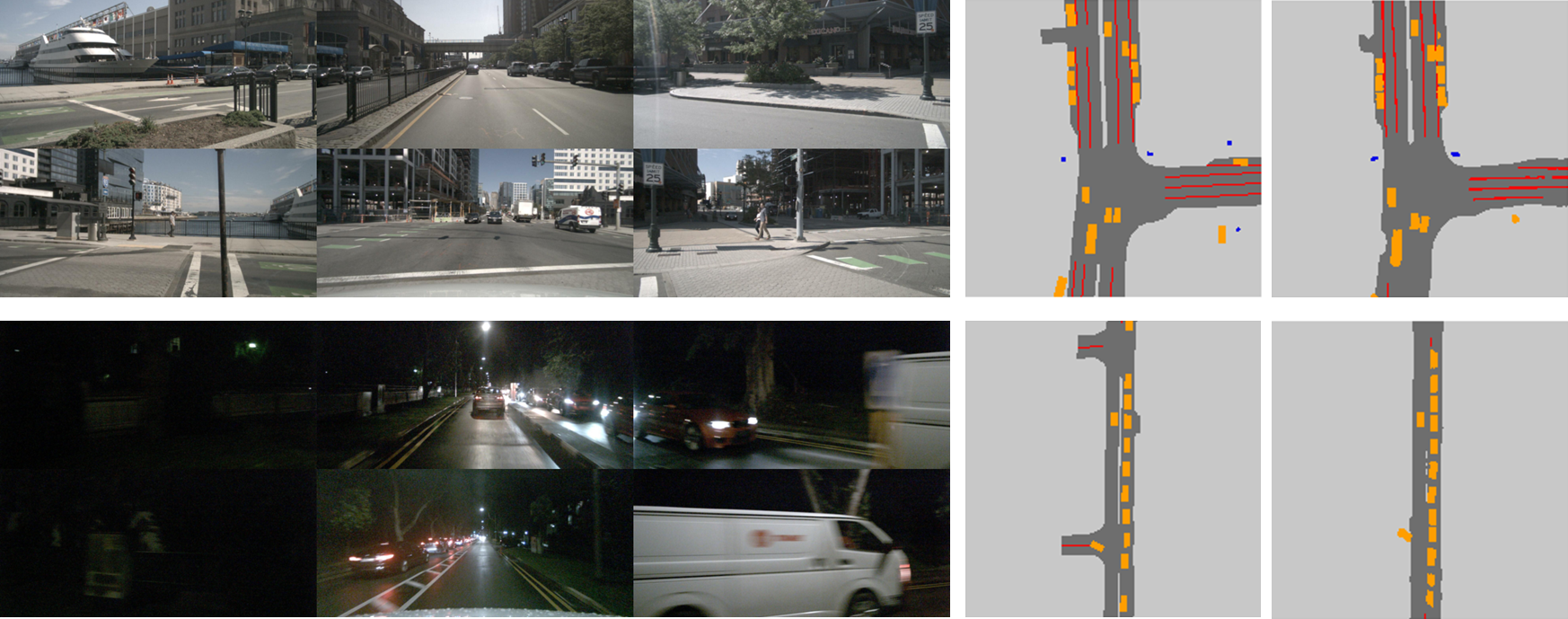}
\caption{Visualization of the prediction results. The first, second, and third columns respectively are the surrounding images, the ground-truth BEV semantic map, and the prediction. \textit{Vehicle}, \textit{Pedestrian}, \textit{Drivable Space}, and \textit{Lane line} are color-coded with orange, blue, grey, and red, respectively.}
\label{fig4}
\end{figure*}
% -----------------------------------------

% Fig 5------------------------------------
\begin{figure}
\centering
\includegraphics[height=5.0cm]{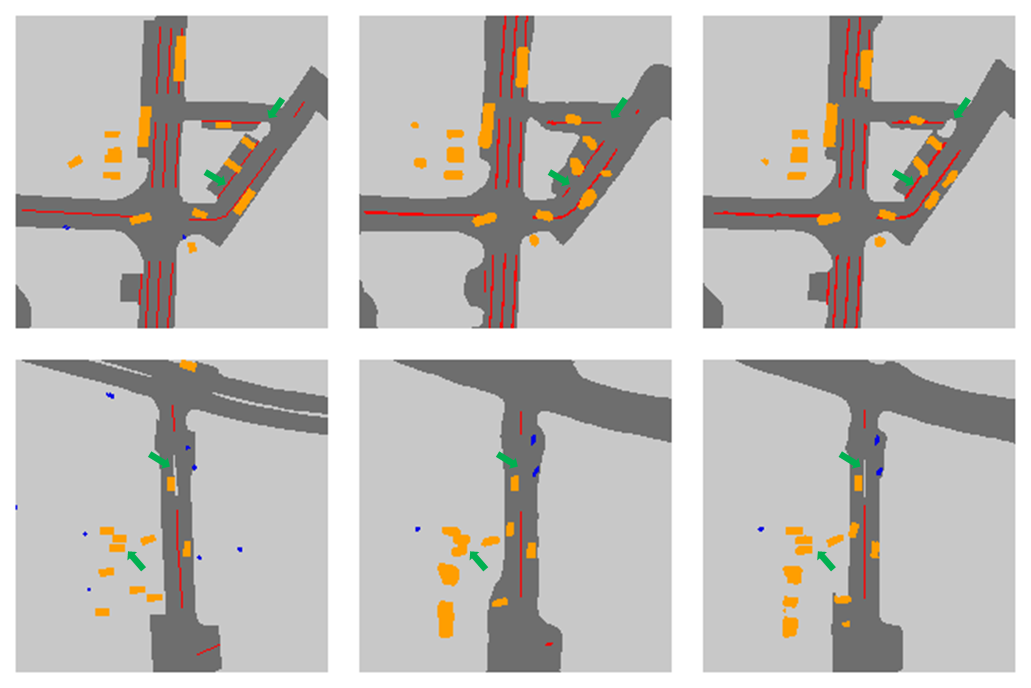}
\caption{Visualization of the ground-truth BEV map and its predictions. The first, second, and third columns respectively correspond to the ground-truth, the prediction from the lowest resolution query map, and the final prediction. The green arrows highlight how missing details are restored through our progressive refinement framework.}
\label{fig5}
\end{figure}
% -----------------------------------------

\section{Experiments}
\subsection{Dataset and Evaluation Settings}
A large-scale real-world dataset, nuScenes \cite{Caesar}, is used to evaluate our model. nuScenes is a collection of 1000 scenes acquired over diverse weather, time-of-day, and traffic conditions. Following the previous works \cite{Li_eccv22, Zhou_cvpr22, Fang_cvpr23}, 28,130 and 6,019 key frames are used for the training and test sets, respectively. To generate the ground-truth BEV semantic maps at a resolution of $(200 \times 200)$, 3D bounding boxes of vehicles and pedestrians or HD map elements in $100m \times 100m$ area around the ego-vehicle are orthographically projected onto the ground plane, following the standard practice \cite{Philion_eccv20, Hu_iccv21}. For an evaluation metric, we use Intersection-over-Union (IoU) score between the ground-truth and its prediction.

\subsection{Network and Training Details}
We use as an image backbone ResNet-50 \cite{He_cvpr16} pre-trained on ImageNet \cite{Russakovsky_arxiv14}. Three consecutive mid-layer feature maps ($N_{l}=3$), each of which corresponds to down-scaling factors of $\times 4$, $\times 8$, and $\times 16$ from the original size, are used as the visual feature maps. The feature maps further undergo a shallow CNN to match the channel dimension with that of the query maps. For the class headers, we use shallow CNNs consisting of Conv, BN \cite{Ioffe_icml15}, and Relu. For the map decoders, we deploy the mask decoder \cite{Li_cvpr22}. Our model is trained with a batch size of 4 for 28 epochs. We optimize our model using AdamW \cite{Loshchilov_iclr21} optimizer with learning rate $2e^{-4}$ and weight decay $1e^{-7}$. We also set $N_{s}=3$ and $\alpha=1$ in Eqn. \ref{eqn5}. Lastly, in accordance with common practice \cite{Zhou_cvpr22, Pan_cvpr23}, we trained a separate NN for each class to achieve the results presented in the subsequent sections. This separation is adopted due to the potential underperformance of jointly trained models, a phenomenon known as \textit{negative transfer} in multi-task learning \cite{Li_eccv22}.

% \textcolor{red}{Finally, following the common practice \cite{Zhou_cvpr22, Pan_cvpr23}, we trained a separate NN for each class to obtain the results shown in the following sections. This is because the jointly trained model may not perform as well as individually trained models, which is common phenomenon called \textit{negative transfer} in multi-task learning \cite{Li_eccv22}}.

\subsection{Subjective and Objective Results}
We objectively compare our model with the SOTA models in Table \ref{table1}. Since our model does not consider previously obtained image data for the current prediction, we report the prediction performance of the $\textit{static}$ version of the SOTA models in the table for fair comparisons. We can see in the table that our proposed model achieves the best performance across almost all class categories. Specifically, our model outperforms BEVFormer-s \cite{Li_eccv22} despite the input image size for our model being four times smaller than that for BEVFormer-s \cite{Li_eccv22}. This result demonstrates that our model effectively captures objects of various sizes in driving scenes. TBPFormer-s \cite{Fang_cvpr23} also shows the performance comparable to ours. However, it has $1.6$ times more trainable parameters than ours ($117.9 \textbf{M}$ v.s. $73.7 \textbf{M}$), which mainly originates from its repeated deployments of SwinT \cite{Liu_iccv21} for the hierarchical query refinement. 

In Figure 4, we visualize the prediction results of our model alongside the ground-truth. We can see in the figure that our model captures the shape of the dynamic road agents and static road elements well. Specifically, our model effectively distinguishes vehicles gathered owing to our query refinement framework and its dedicated training method. Figure \ref{fig5} further demonstrates the effectiveness of the framework.

% ------------------------------------------------------------ Table 2
\begin{table}[t]
\begin{center}
\caption{Ablation study conducted on nuScenes}\label{tabletwo}
\begin{tabular}{|c|c|c|}
\hline
Model & Vehicle & Drivable Space \\
\hline
\textbf{M1} & 42.1 & 79.6 \\ % no VFI
\textbf{M2} & 42.3 & 79.8 \\ % VFI w/o Inter-CIM
\textbf{M3} & 42.8 & 80.3 \\ % VFI with conventional deform attn
\hline
\textbf{M4} & 42.0 & 79.0 \\ % w/o progressive refinement, 
\textbf{M5} & 43.3 & 80.3 \\ % w/ progressive refinement & w/o auxilary task
\textbf{M6} & 43.2 & 81.0 \\ % w/ progressive refinement & w/o skip-connect & all auxiliary
\hline
\hline
\textbf{Ours} & \textbf{43.7} & \textbf{81.1} \\ % baseline
\hline
\end{tabular}
\end{center}
\end{table}
% ------------------------------------------------------------

\subsection{Ablation Study}
In Table \ref{tabletwo}, we demonstrate the effectiveness of the VFI module and our progressive refinement framework in terms of the prediction performance. In the table, $\textbf{M1}$ and $\textbf{M2}$ denote the proposed model without the VFI module and with the Inter-CIM removed, respectively. For $\textbf{M3}$, we replace the proposed deformable attention for the Inter-CIM with the conventional one. The table shows that the proposed deformable attention enhances the interaction between the features across images, leading to a significant improvement in prediction performance compared to the conventional attention (\textbf{M3} v.s. \textbf{Ours}). $\textbf{M4}$ and $\textbf{M5}$ respectively denote the proposed model with a single query map ($N_{s}=1$) and MR query maps ($N_{s}=3$) without applying the auxiliary task during the training. In contrast, for $\textbf{M6}$ ($N_{s}=3$), we exclude the addition of the lowest resolution map to the highest one and apply the auxiliary task for all query maps except for the final resolution. The table and Figure \ref{fig5} demonstrate that the proposed progressive query refinement framework, along with its dedicated training (the auxiliary task), enhances our network's ability to capture the global and local characteristic of driving scenes.

\section{CONCLUSIONS}
This paper proposes a new NN architecture that progressively updates and merges query maps from the lowest resolution to the final resolution for BEV semantic segmentation. Our model's ability to capture the global and local characteristics of driving scenes is significantly improved through the proposed residual learning. In addition, a new visual feature interaction method is proposed to further enhance the prediction performance of our model. Our future research plan includes incorporating temporal information such as previously obtained image data into our framework.

% In addition, the interaction between visual features across feature levels and images is promoted by the proposed VFI module, significantly contributing to performance improvement. Our experiments demonstrate that our proposed model outperforms the SOTA models in terms of the prediction accuracy. Our future research plan includes incorporating temporal information such as previously obtained image data into our framework.

\addtolength{\textheight}{-12cm}   % This command serves to balance the column lengths

%%%%%%%%%%%%%%%%%%%%%%%%%%%%%%%%%%%%%%%%%%%%%%%%%%%%%%%%%%%%%%%%%%%%%%%%%%%%%%%%

%%%%%%%%%%%%%%%%%%%%%%%%%%%%%%%%%%%%%%%%%%%%%%%%%%%%%%%%%%%%%%%%%%%%%%%%%%%%%%%%

%%%%%%%%%%%%%%%%%%%%%%%%%%%%%%%%%%%%%%%%%%%%%%%%%%%%%%%%%%%%%%%%%%%%%%%%%%%%%%%%
%\section*{APPENDIX}


%\section*{ACKNOWLEDGMENT}
%This work was supported by IITP grant funded by the Korea government (MSIT) (RS-2023-00236245, Development of Perception/Planning AI SW for Seamless Autonomous Driving in Adverse Weather/Unstructured Environment)

%%%%%%%%%%%%%%%%%%%%%%%%%%%%%%%%%%%%%%%%%%%%%%%%%%%%%%%%%%%%%%%%%%%%%%%%%%%%%%%%

% References are important to the reader; therefore, each citation must be complete and correct. If at all possible, references should be commonly available publications.

\bibliography{IEEEexample}
\bibliographystyle{IEEEtranS}

\end{document}